\documentclass[letterpaper]{article} 
\usepackage{aaai25}  
\usepackage{times}  
\usepackage{helvet}  
\usepackage{courier}  
\usepackage[hyphens]{url}  
\usepackage{graphicx} 
\usepackage{natbib}  
\usepackage{caption} 
\usepackage{booktabs}
\usepackage{subcaption}
\usepackage{algorithm}
\usepackage{algorithmic}
\usepackage{newfloat}
\usepackage{listings}
\DeclareCaptionStyle{ruled}{labelfont=normalfont,labelsep=colon,strut=off} 
\floatstyle{ruled}
\newfloat{listing}{tb}{lst}{}
\floatname{listing}{Listing}
\usepackage{fancyhdr}
\fancypagestyle{firstpagehf}
{
    \fancyhf{}
    \fancyhead[HC]{The AAAI 2025 Workshop on Multi-Agent AI in the Real World (MARW)}
    \fancyfoot[C]{\thepage}
}
\nocopyright
\title{Multi-Agent Collaboration in Incident Response with Large Language Models}
\author {
    Zefang Liu
}
\affiliations{
    Georgia Institute of Technology\\
    North Avenue, Atlanta, GA 30332 USA\\
    liuzefang@gatech.edu
}
\begin{document}
\thispagestyle{firstpagehf}
\maketitle
\begin{abstract}
Incident response (IR) is a critical aspect of cybersecurity, requiring rapid decision-making and coordinated efforts to address cyberattacks effectively. Leveraging large language models (LLMs) as intelligent agents offers a novel approach to enhancing collaboration and efficiency in IR scenarios. This paper explores the application of LLM-based multi-agent collaboration using the Backdoors \& Breaches framework, a tabletop game designed for cybersecurity training. We simulate real-world IR dynamics through various team structures, including centralized, decentralized, and hybrid configurations. By analyzing agent interactions and performance across these setups, we provide insights into optimizing multi-agent collaboration for incident response. Our findings highlight the potential of LLMs to enhance decision-making, improve adaptability, and streamline IR processes, paving the way for more effective and coordinated responses to cyber threats.
\end{abstract}
\section{Introduction}

Effective incident response (IR) \citep{kruse2001computer,luttgens2014incident} is critical in mitigating the impact of cyberattacks on organizations. IR requires timely decision-making, coordination across roles, and rapid adaptation to evolving threats. Traditional incident response teams rely on human expertise, collaboration, and structured processes to identify and address attack vectors. However, as cyber threats grow in complexity and frequency, augmenting human teams with intelligent systems becomes increasingly appealing.

Large language models (LLMs) \citep{naveed2023comprehensive} have demonstrated exceptional potential in tasks involving natural language understanding, reasoning, and decision-making. Beyond their general-purpose applications, LLMs enable new opportunities for multi-agent collaboration \citep{guo2024large}, facilitating communication, strategy, and problem-solving among diverse roles. Their applications span a wide range of fields, including society simulation \citep{park2023generative}, mechanical problem-solving \citep{ni2024mechagents}, healthcare \citep{wang2024colacare}, and supply chain management \citep{quan2024invagent}. In the field of cybersecurity, LLMs are increasingly being utilized for benchmarking \citep{liu2023secqa,liu2024cyberbench,tihanyi2024cybermetric} and practical applications \citep{motlagh2024large,xu2024large,zhang2024llms}. Within incident response, LLMs can simulate human-like agents to coordinate investigations, identify attack patterns, and recommend effective countermeasures \citep{hays2024employing}.

\begin{figure}[t!]
    \centering
    \includegraphics[width=.95\columnwidth]{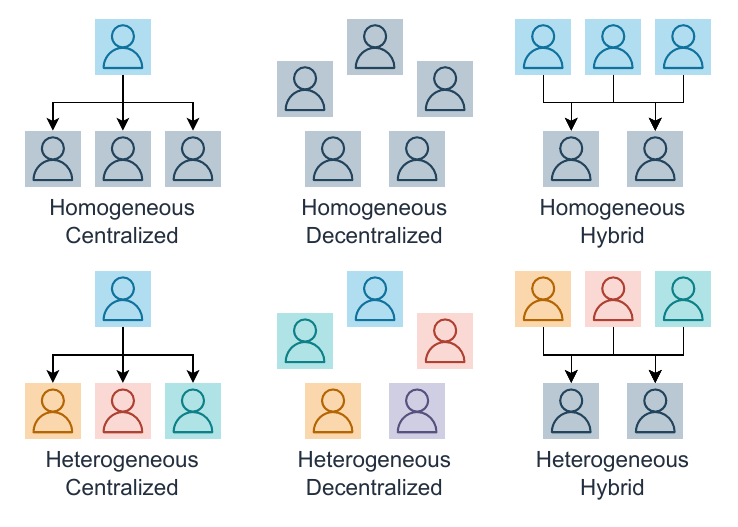}
    \caption{Visualization of defender team structures used in the Backdoors \& Breaches incident response simulation.}
    \label{fig:team-structures}
\end{figure}

This paper\footnote{GitHub repository: \url{https://github.com/zefang-liu/AutoBnB}} explores how LLM-based multi-agent systems perform in simulated incident response scenarios, using the Backdoors \& Breaches \citep{backdoorsandbreaches} tabletop game as a testbed. The game serves as a structured environment where agents must collaborate to uncover hidden attack vectors through investigative procedures and strategic planning \citep{young2021backdoors}. As shown in Figure \ref{fig:team-structures}, we examine various team structures-centralized, decentralized, and hybrid-to understand how role design and communication dynamics influence team effectiveness. Through a series of experiments, we evaluate the performance of LLM-based agents in detecting and mitigating simulated cyberattacks. Our findings reveal insights into the strengths and challenges of multi-agent systems for incident response, providing a foundation for future advancements in cybersecurity collaboration.

\section{Background}

In this section, we provide an overview of incident response as a critical cybersecurity process and discuss the role of large language model (LLM)-based multi-agent systems in facilitating collaboration and decision-making.

\subsection{Incident Response}

Incident response (IR) \citep{schlette2021comparative} is a structured approach to managing and mitigating the impact of cybersecurity incidents. It involves the identification, containment, eradication, and recovery from security breaches to minimize damage and restore normal operations \citep{cichonski2012computer}. Effective IR necessitates collaboration among various stakeholders, including information technology (IT) professionals, security analysts, and organizational leadership, to ensure a coordinated and timely response. The complexity and sophistication of modern cyber threats have heightened the need for efficient and adaptive incident response strategies. Traditional IR processes often rely on manual interventions and predefined protocols, which may not suffice against rapidly evolving attack vectors. Consequently, there is a growing interest in integrating advanced technologies, such as machine learning and language models \citep{shaukat2020performance,nilua2020machine,macas2022survey,liu2024review,hays2024employing}, to enhance the agility and effectiveness of incident response efforts.

\subsection{LLM-Based Multi-Agent Systems}

Large language models (LLMs) \citep{kaddour2023challenges} have demonstrated remarkable capabilities in natural language understanding and generation, enabling them to perform complex reasoning and decision-making tasks. Leveraging these capabilities, LLM-based multi-agent systems have emerged as a promising approach to simulate and enhance collaborative problem-solving in various domains. Recent studies \citep{talebirad2023multi,li2023theory,guo2024large,wang2024survey} have explored the integration of LLMs into multi-agent frameworks, highlighting their potential to improve coordination, efficiency, and applicability in complex problem-solving and world simulation. In the context of cybersecurity, LLM-based agents can emulate the roles of human defenders, enabling automated analysis, detection, and response to security incidents. These systems hold significant potential to streamline and enhance incident response processes, addressing the inefficiencies and constraints of traditional methods.

\section{Incident Response Game}

Backdoors \& Breaches (B\&B) is a tabletop card game designed to simulate the complexities of cybersecurity incident response in a collaborative and engaging manner. Developed by \citet{backdoorsandbreaches}, the game offers a hands-on approach to learning key strategies for identifying, containing, and mitigating cyber threats. By combining structured gameplay with elements of randomness, B\&B immerses players in realistic decision-making processes, fostering critical thinking and teamwork. A detailed discussion on the relation between B\&B and real-world cybersecurity incident response is provided in Appendix \ref{sec:relation}.

\subsection{Game Objective and Setup}

The primary objective of B\&B is for defenders to uncover and reveal four hidden attack cards within a maximum of 10 turns. These attack cards represent the stages of a cyberattack: initial compromise, pivot and escalate, command and control (C2) and exfiltration, and persistence.

At the beginning of the game, the incident captain sets the stage by selecting one hidden attack card for each phase. The defenders, working collaboratively, must employ investigative procedures to reveal these cards. Success is determined by strategic selection of procedure cards, team coordination, and adaptive responses to in-game events.

\subsection{Card Types}

\begin{figure}[h!]
    \centering
    \begin{subfigure}[b]{0.42\columnwidth}
        \centering
        \includegraphics[width=\textwidth]{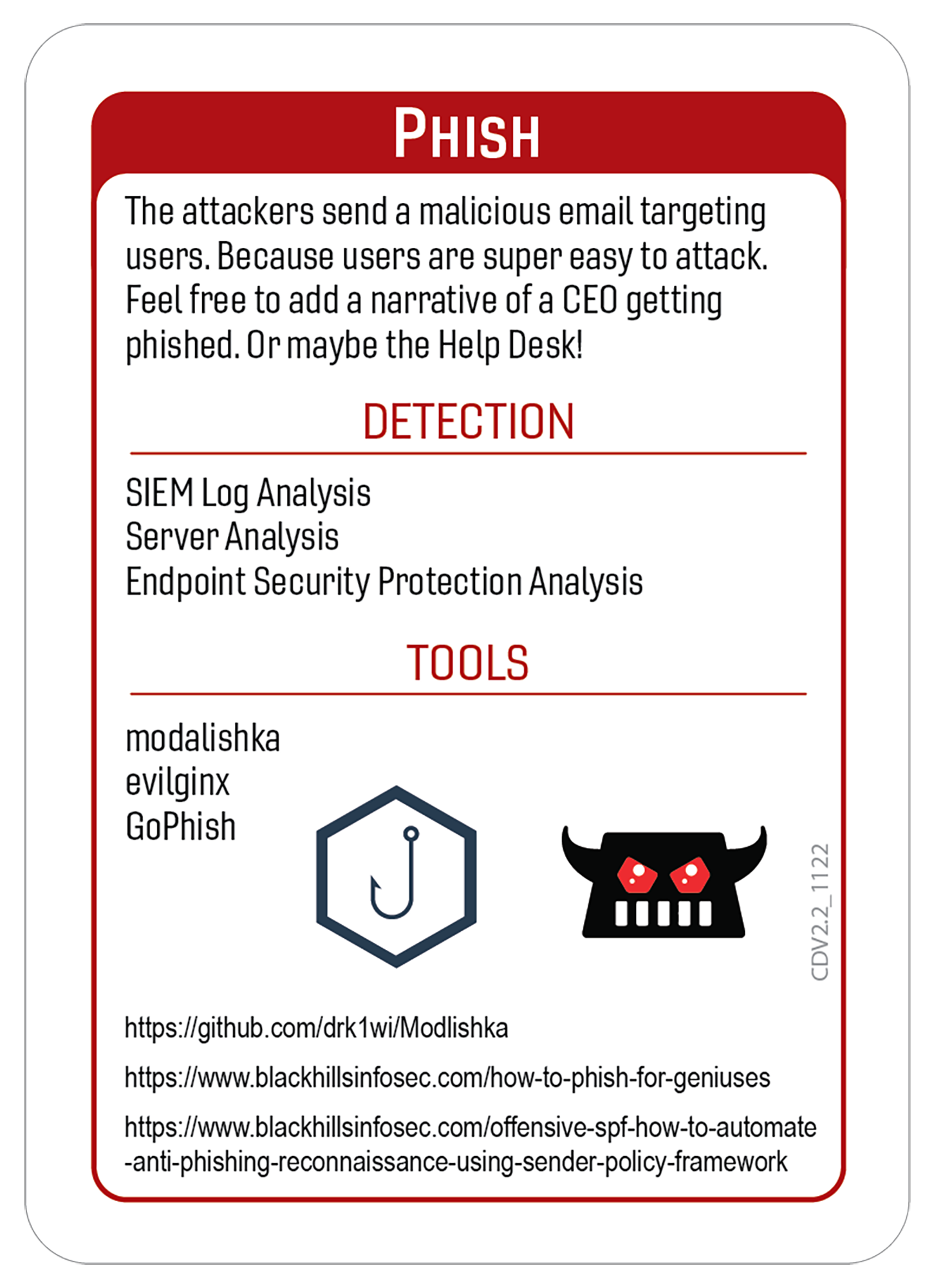}
        \caption{Initial compromise card}
    \end{subfigure}
    \hfill
    \begin{subfigure}[b]{0.42\columnwidth}
        \centering
        \includegraphics[width=\textwidth]{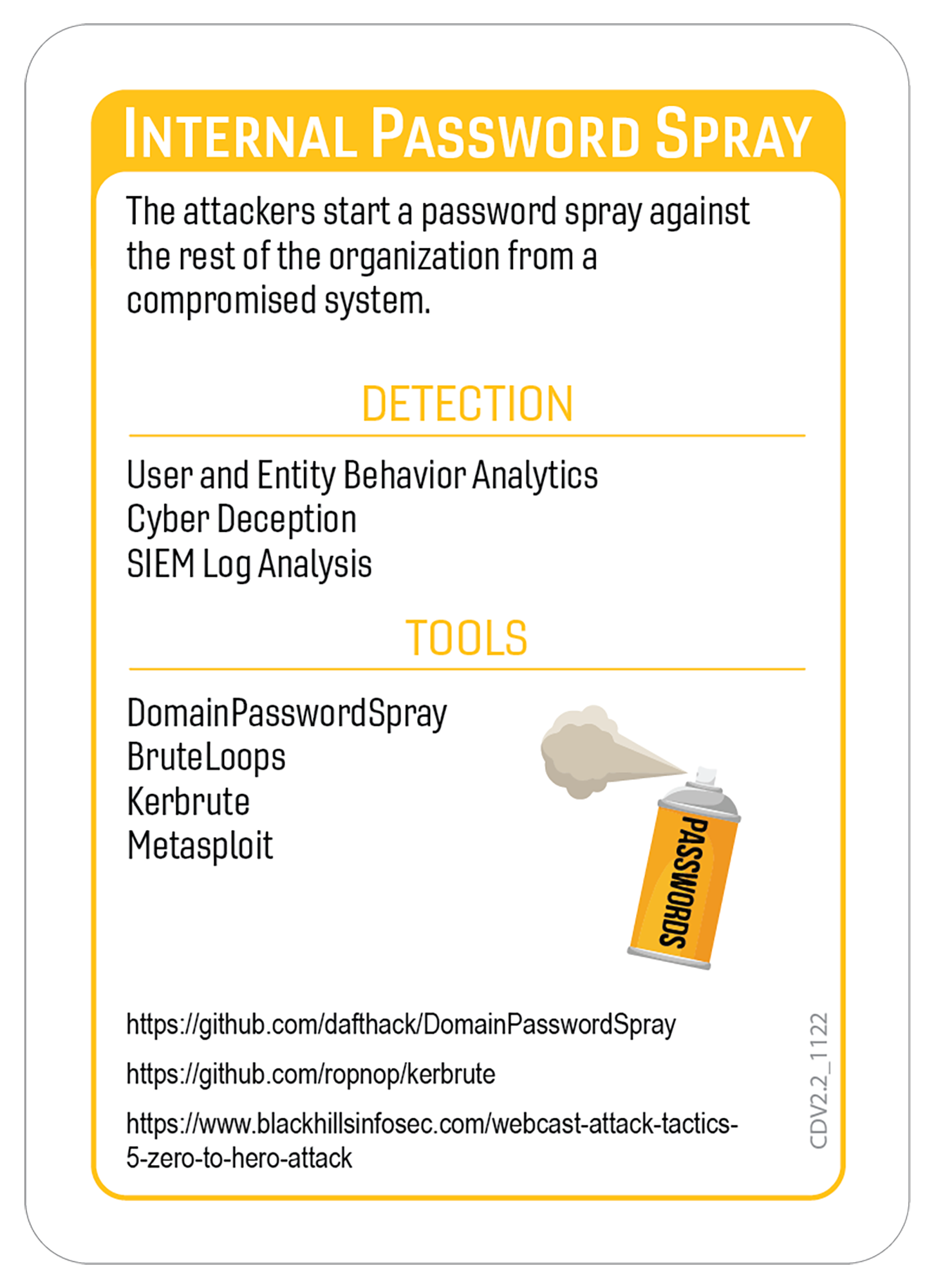}
        \caption{Pivot and escalate card}
    \end{subfigure}
    \begin{subfigure}[b]{0.42\columnwidth}
        \centering
        \includegraphics[width=\textwidth]{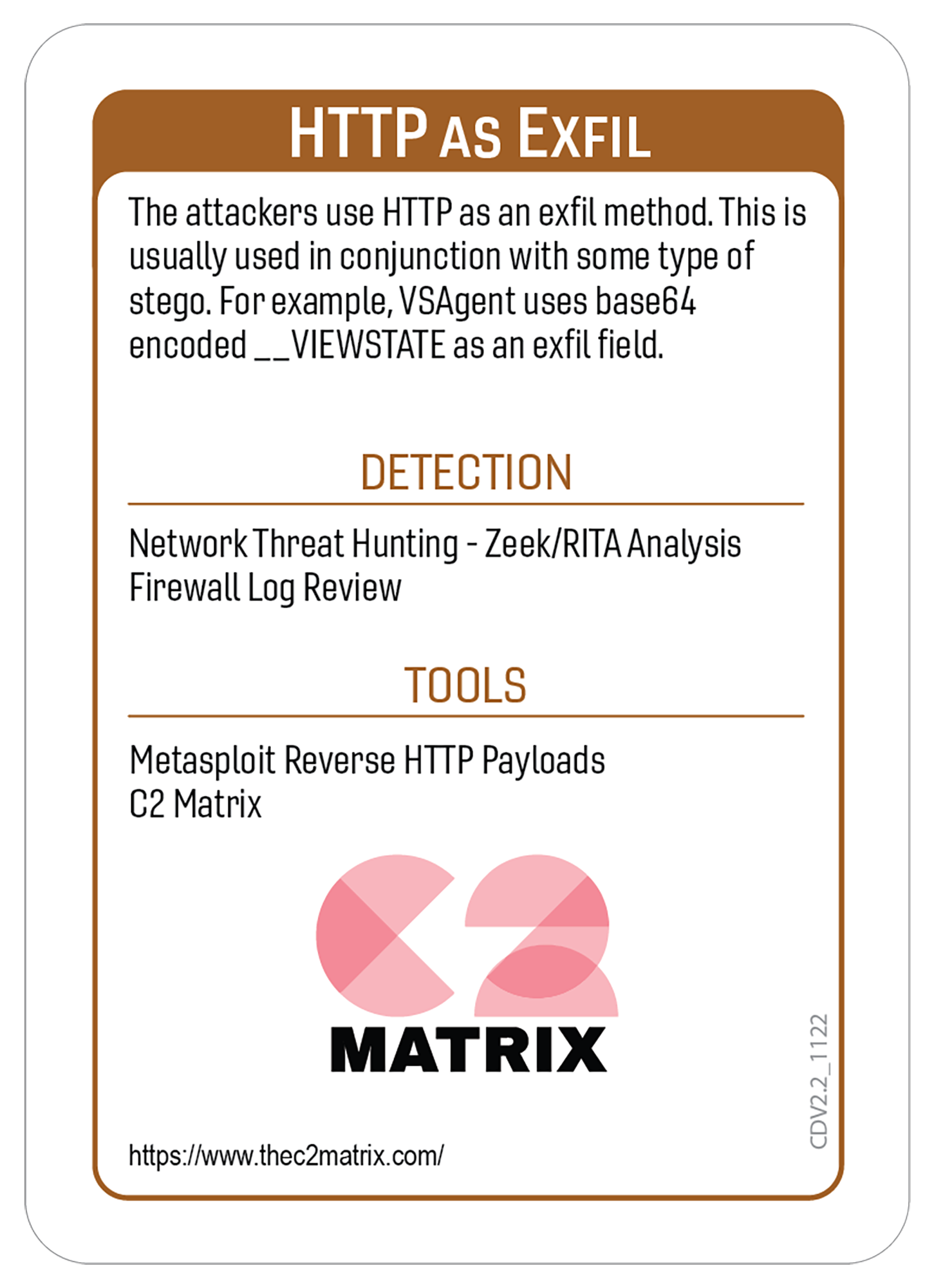}
        \caption{C2 and exfil card}
    \end{subfigure}
    \hfill
    \begin{subfigure}[b]{0.42\columnwidth}
        \centering
        \includegraphics[width=\textwidth]{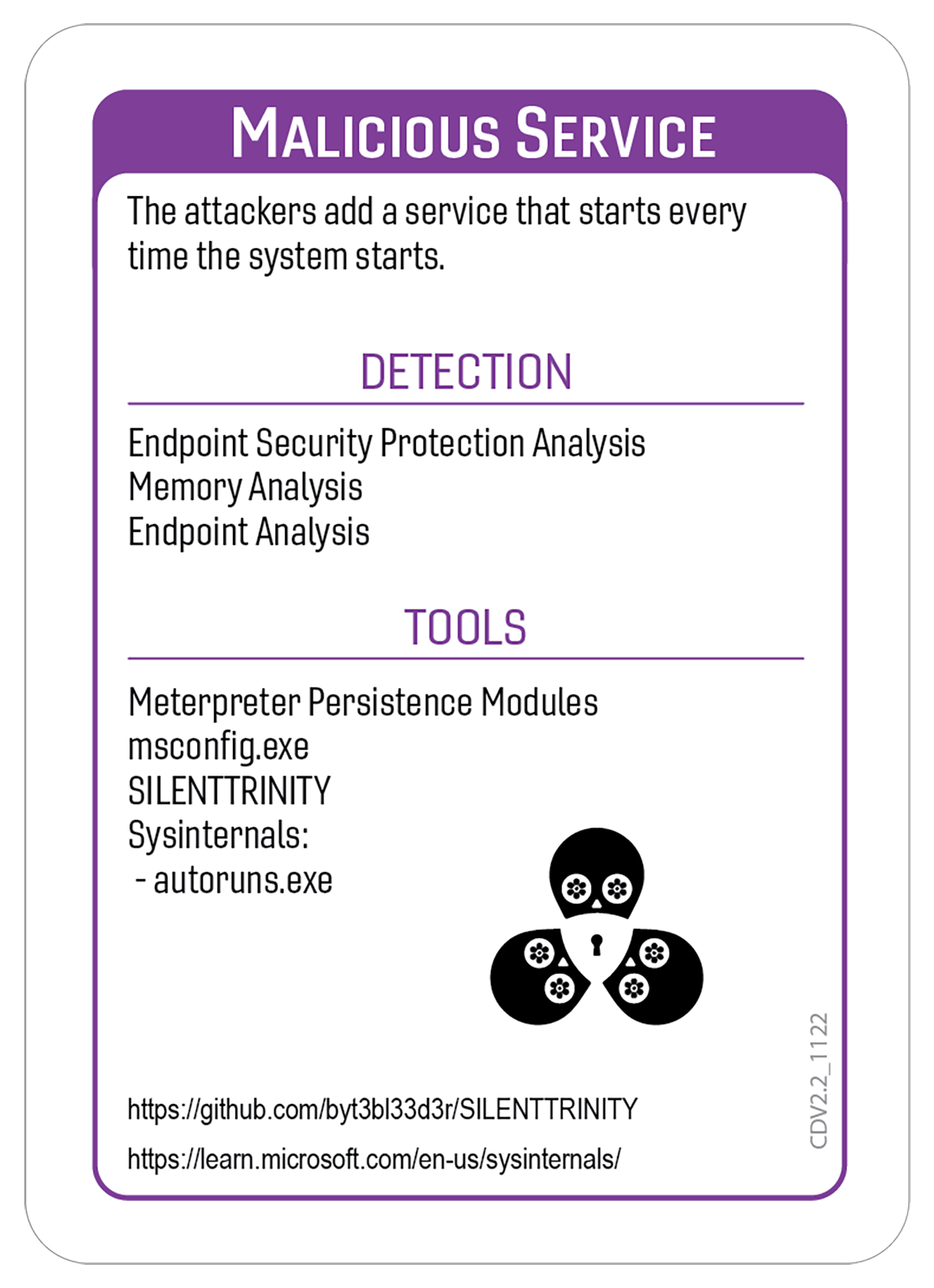}
        \caption{Persistence card}
    \end{subfigure}
    \begin{subfigure}[b]{0.42\columnwidth}
        \centering
        \includegraphics[width=\textwidth]{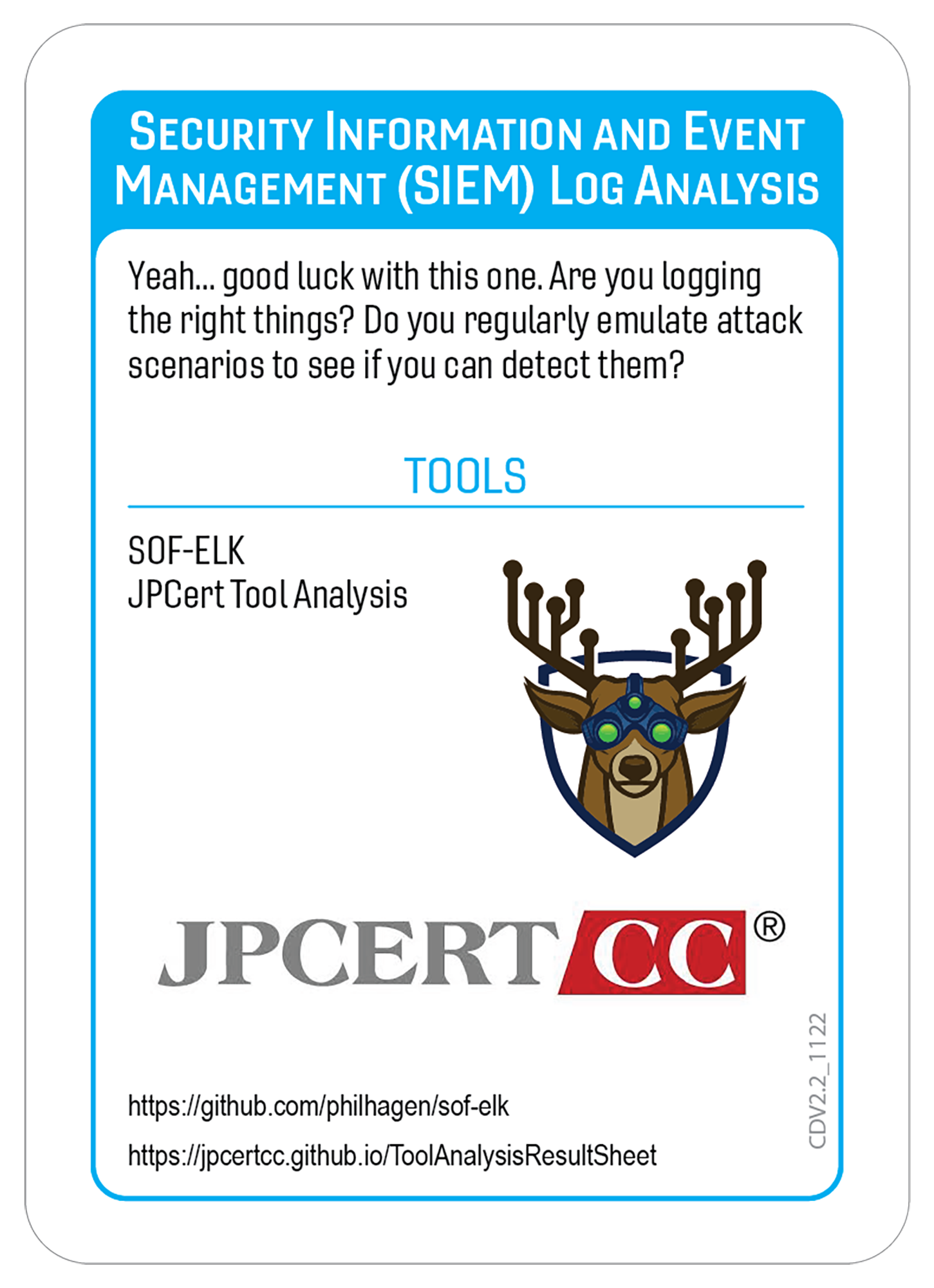}
        \caption{Procedure card}
    \end{subfigure}
    \hfill
    \begin{subfigure}[b]{0.42\columnwidth}
        \centering
        \includegraphics[width=\textwidth]{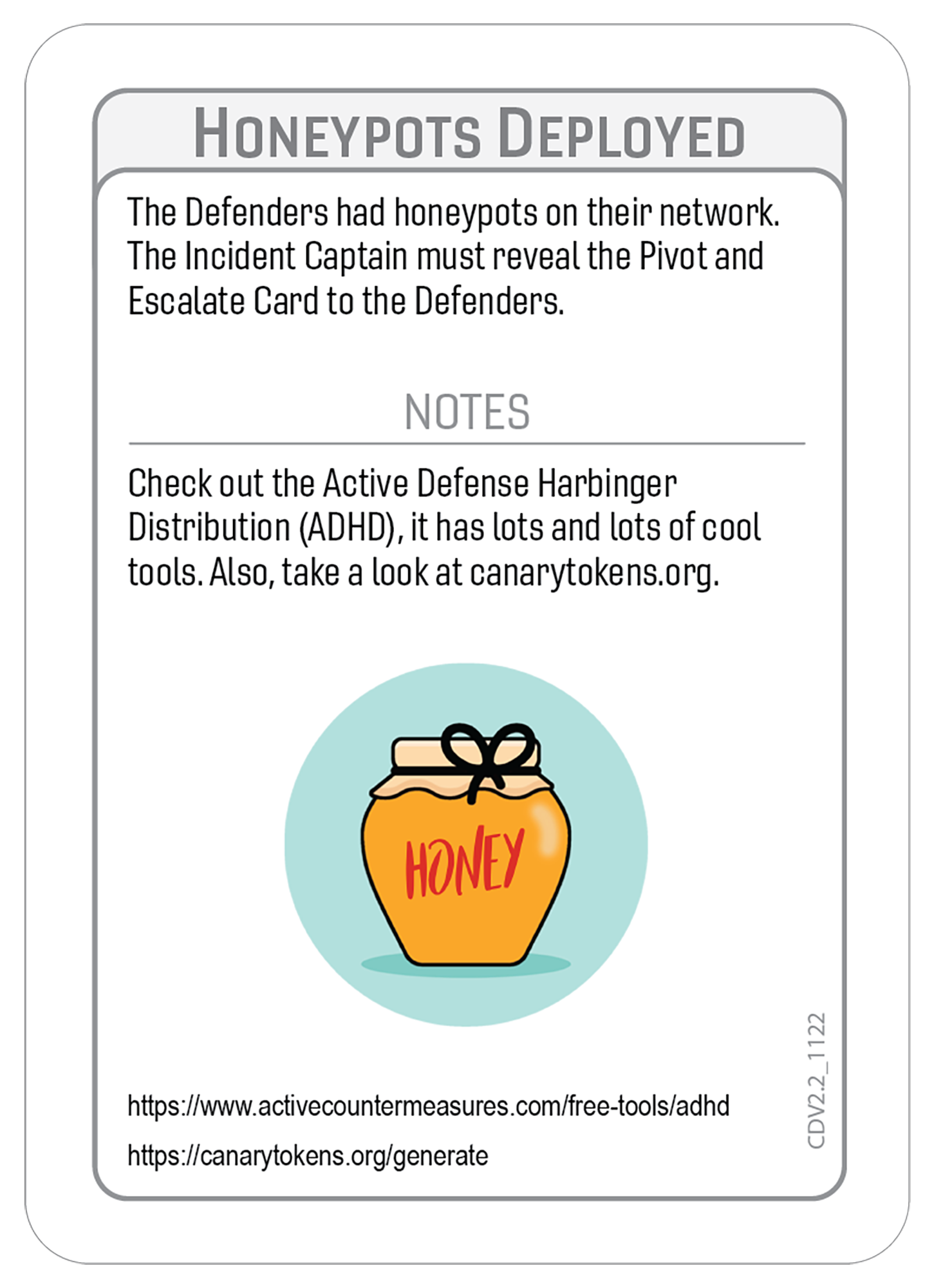}
        \caption{Inject card}
    \end{subfigure}
    \caption{Examples of Backdoors \& Breaches cards used in this study. Image source: Black Hills Information Security.}
    \label{fig:bnb-cards}
\end{figure}

The gameplay centers around three distinct card types (Figure \ref{fig:bnb-cards}), each simulating a key aspect of incident response:
\begin{itemize}
    \item \textbf{Attack cards (32 cards):} Represent adversarial tactics and techniques during a cyberattack, tied to specific detection methods that defenders must identify and execute.

    \item \textbf{Procedure cards (11 cards):} Tools and investigative strategies defenders use to uncover attack cards. They are divided into established procedures, offering a +3 dice roll modifier due to their reliability, and other procedures with no modifier. Each card enters a three-turn cooldown after use, reflecting operational constraints.

    \item \textbf{Inject cards (9 cards):} Introduce unpredictable events or challenges, mirroring the uncertainties of real-world incident response. They are triggered by specific conditions, such as a natural dice roll of 1 or 20 or three consecutive failed attempts.
\end{itemize}
A complete list of B\&B cards can be found in Appendix \ref{sec:bnb-cards}.

\subsection{Game Mechanics}

The structured gameplay of B\&B emphasizes strategic decision-making and adaptability. Each turn, the defenders select a procedure card and roll a 20-sided dice to determine the outcome of their attempt. A final roll of 11 or higher, including any applicable modifiers, results in success, while 10 or lower constitutes a failure. Successful attempts reveal hidden attack cards if the procedure matches one of their detection methods. Failures not only hinder progress but also contribute to a consecutive failure count. If three consecutive attempts fail, an inject card is drawn, introducing a new challenge to the game. Similarly, natural dice rolls of 1 or 20 can also trigger inject events, adding an element of randomness and unpredictability.

\subsection{Roles and Collaboration}

B\&B is inherently collaborative, requiring players to take on distinct roles to ensure an effective response. The incident captain oversees the game, sets the scenario, and introduces inject events as required. The defenders, on the other hand, are tasked with selecting and executing the appropriate procedures to counteract the hidden attack cards. Depending on the game configuration, defenders may include generalists and domain-specific experts, such as specialists in endpoint security, network traffic analysis, or log analytics.

\section{Methodology}

This section outlines our study on LLM-driven multi-agent collaboration for incident response, using Backdoors \& Breaches (B\&B) as a framework where LLMs assumed roles, coordinated decisions, and adapted to dynamic scenarios.

\subsection{Study Design}

This study aims to evaluate the effectiveness of LLM-based multi-agent collaboration in identifying hidden attack cards within simulated incident response scenarios. The experiment simulates realistic cybersecurity incident response dynamics using predefined agent roles, including an incident captain and multiple defenders with their system messages in Appendix \ref{sec:sequence-of-game} and \ref{sec:system-messages}. The incident captain agent is responsible for setting up scenarios with hidden attack cards representing various breach stages, introducing procedure cards, managing inject events, and ensuring adherence to game rules. To encourage team decision-making and collaboration, the captain facilitates gameplay without revealing correct solutions.

Defender agents represent team members with varying expertise and responsibilities, reflecting predefined roles such as endpoint security expert or network traffic analysis expert. These agents communicate through a group chat, collaboratively selecting and executing procedure cards to detect and mitigate attack stages. The study examines the interactions and decision-making processes within different team structures, including centralized, decentralized, and hybrid configurations, each offering unique dynamics in leadership, role specialization, and collaboration.

The simulation is implemented using the AutoGen \cite{wu2023autogen} framework, which enables seamless interaction among agents through a shared group chat managed by a group chat manager. The agents autonomously communicate, coordinate decisions, and adapt to game events such as dice outcomes and inject challenges. A dedicated tool executor manages backend operations, including card drawing, dice rolling, and data logging, ensuring consistency and accuracy in procedural execution. By combining structured roles, automated tools, and systematic data collection, this study offers valuable insights into the impact of team structure on multi-agent collaboration and decision-making in incident response scenarios.

\subsection{Team Structures}

To evaluate the impact of organizational collaboration on incident response, we employed six distinct team structures (Figure \ref{fig:team-structures}) that reflect common approaches to teamwork in cybersecurity. These configurations allowed us to analyze how decision-making, role specialization, and leadership influence team performance within the B\&B simulation:

\begin{itemize}
    \item \textbf{Homogeneous centralized structure:} A team leader directed a group of generalist defenders, emphasizing hierarchical decision-making.
    
    \item \textbf{Heterogeneous centralized structure:} A team leader coordinated defenders with expertise in distinct domains, highlighting specialized collaboration under centralized leadership.
    
    \item \textbf{Homogeneous decentralized structure:} Generalist defenders operated collaboratively without a designated leader, relying on consensus-based decision-making. 
    
    \item \textbf{Heterogeneous decentralized structure:} Specialists from distinct domains worked autonomously, contributing equally to team decisions without centralized leadership.
    
    \item \textbf{Homogeneous hybrid structure:} A mixed team of generalist experts and beginners collaborated, with experts guiding decision-making and mentoring the beginners.
    
    \item \textbf{Heterogeneous hybrid structure:} Specialists in distinct domains collaborated with beginners, combining expert knowledge with opportunities for skill development.
\end{itemize}

These team structures were designed to simulate realistic dynamics in cybersecurity operations, enabling a comprehensive analysis of their strengths, limitations, and applicability to real-world incident response scenarios.

\section{Experiments}

In this section, we evaluate the effectiveness of LLM-based multi-agent collaboration in incident response by conducting a series of simulations using the Backdoors \& Breaches (B\&B) framework under various team structures.

\subsection{Experimental Setup}

We evaluate the performance of LLM-based multi-agent collaboration in incident response by simulating six distinct team structures within the B\&B game using the AutoGen framework \citep{wu2023autogen}. Each agent, powered by GPT-4o \citep{achiam2023gpt} and configured with a temperature of 1, is assigned a predefined role with tailored responsibilities. The team structures include five defenders, varying in leadership, expertise, and collaboration styles: \textbf{homogeneous centralized structure:} 1 team leader and 4 team members; \textbf{heterogeneous centralized structure:} 1 team leader, 1 endpoint security expert, 1 network traffic analysis expert, 1 log and behavioral analysis expert, and 1 deception and containment expert; \textbf{homogeneous decentralized structure:} 5 team members; \textbf{heterogeneous decentralized structure:} 1 endpoint security expert, 1 network traffic analysis expert, 1 log and behavioral analysis expert, 1 deception and containment expert, and 1 incident response expert; \textbf{homogeneous hybrid structure:} 3 experts and 2 beginners; \textbf{heterogeneous hybrid structure:} 1 endpoint security expert, 1 network traffic analysis expert, 1 log and behavioral analysis expert, and 2 beginners. We conduct 20 simulations for each team structure, with one example of a game simulation provided in Appendix \ref{sec:game-example} for the homogeneous centralized team structure.

\subsection{Experimental Results}

Table \ref{tab:results} summarizes the performance outcomes for different team structures in simulated incident response scenarios, including the number of successes, failures, penetration tests, and invalid results (e.g., prematurely ending scenarios). Homogeneous centralized and hybrid structures achieved the highest success rates with 14 successful simulations each, showcasing the benefits of clear leadership and streamlined communication. Hybrid structures also fostered collaboration between experts and beginners, accelerating teamwork through mentorship and shared decision-making. Homogeneous decentralized teams performed well, achieving 13 successes, likely due to simplified coordination among generalists. In contrast, heterogeneous structures, both decentralized and centralized, faced challenges in reaching consensus due to domain experts' differing perspectives, particularly without leadership. These findings highlight the need to balance team composition, leadership, and collaboration strategies to optimize incident response in multi-agent systems.

\begin{table}[h!]
\centering
\begin{tabular}{lrrrr}
\toprule
\textbf{Team} & \textbf{Success} & \textbf{Failure} & \textbf{Pentest} & \textbf{Invalid} \\
\midrule
Homo-Cen & 14 & 1 & 2 & 3 \\
Heter-Cen & 13 & 3 & 3 & 1 \\
Homo-Dec & 13 & 1 & 1 & 5 \\
Hetero-Dec & 12 & 3 & 3 & 2 \\
Homo-Hyb & 14 & 3 & 2 & 1 \\
Hetero-Hyb & 13 & 1 & 2 & 4 \\
\bottomrule
\end{tabular}
\caption{Performance outcomes across different team structures in incident response simulations.}
\label{tab:results}
\end{table}

\subsection{Case Studies}

We analyzed 12 failure cases across different team structures to identify key challenges that impeded effective incident response, with detailed analyses provided in Appendix \ref{sec:case-study-details}. Homogeneous centralized teams often over-relied on standard procedures, failing to adapt to subtle breach indicators. Heterogeneous centralized teams struggled with prioritization, neglecting critical network-level indicators or delaying behavior-based analytics. Decentralized structures, both homogeneous and heterogeneous, frequently relied excessively on high-modifier procedures while neglecting areas like behavior analytics. Hybrid teams revealed the importance of aligning diverse expertise, with failures stemming from misaligned priorities or underutilized procedures like memory and log analysis. These findings highlight the need for adaptive and context-aware procedure selection to optimize incident response outcomes across team structures.

\section{Conclusion}

In this study, we explored the potential of large language models (LLMs) to enhance multi-agent collaboration in incident response scenarios, leveraging the Backdoors \& Breaches framework as a structured simulation environment. By analyzing centralized, decentralized, and hybrid team configurations, we demonstrated how LLMs can improve decision-making, adaptability, and collaboration in addressing cybersecurity challenges. The findings emphasize the critical role of team structure in optimizing performance and highlight the value of integrating LLM-based agents to augment incident response strategies. This study paves the way for future research into refining LLM-driven collaboration frameworks to meet the dynamic and complex demands of modern cybersecurity.

\section*{Acknowledgments}

We gratefully acknowledge the OpenAI API credits generously provided by Yinzhu Quan, which significantly contributed to the facilitation of this research.

\bibliography{aaai25}
\newpage
\appendix
\section{Relation to Real Incident Response}
\label{sec:relation}

The Backdoors \& Breaches (B\&B) framework provides a structured and engaging simulation of real-world cybersecurity incident response, emphasizing critical skills such as strategic planning, team collaboration, and adaptive decision-making under uncertainty. By mirroring key stages of a cyberattack—initial compromise, pivot and escalate, command and control, and persistence—the game aligns closely with the dynamics of real-world incident response, where teams must detect threats, prioritize actions, and address unexpected challenges. Within B\&B, the incident captain, analogous to a cybersecurity team lead, crafts scenarios that replicate the complexity and unpredictability of real breaches, while defenders collaboratively apply investigative procedures and respond to inject events representing disruptions or resource constraints. However, it is important to acknowledge the limitations of the framework. The game abstracts critical aspects of real incidents, such as the scale and complexity of infrastructures, the iterative nature of adversarial attacks, and the need for cross-departmental communication during live breaches. Additionally, while randomness through dice rolls and inject cards captures some unpredictability, it cannot fully emulate the evolving tactics of real attackers. Despite these limitations, B\&B remains a valuable tool for practicing the core principles of incident response and serves as a foundation for exploring large language model agent collaboration in realistic cybersecurity scenarios.

\section{Backdoors \& Breaches Cards}
\label{sec:bnb-cards}

In this appendix, we present a categorized list of all Backdoors \& Breaches card types used in this study.

\begin{itemize}
    \item \textbf{Attack Cards:}
    \begin{itemize}
        \item \textbf{Initial Compromise (10 cards):} Phish, Web Server Compromise, External Cloud Access, Insider Threat, Password Spray, Trusted Relationship, Social Engineering, Bring Your Own (Exploited) Device, Exploitable External Service, Credential Stuffing.

        \item \textbf{Pivot and Escalate (7 cards):} Internal Password Spray, Kerberoasting/ASREPRoasting, Broadcast/Multicast Protocol Poisoning, Weaponizing Active Directory, Credential Stuffing, New Service Creation/Modification, Local Privilege Escalation.

        \item \textbf{Command and Control (C2) (6 cards):} HTTP as Exfil, HTTPS as Exfil, DNS as C2, Windows Background Intelligent Transfer Service (BITS), Gmail/Tumblr/Salesforce/Twitter as C2, Domain Fronting as C2.

        \item \textbf{Persistence (9 cards):} Malicious Service, DLL Attacks, Malicious Driver, New User Added, Application Shimming, Malicious Browser Plugins, Logon Scripts, Evil Firmware, Accessibility Features.
    \end{itemize}

    \item \textbf{Procedure Cards (11 cards):} Security Information and Event Management (SIEM) Log Analysis, Server Analysis, Firewall Log Review, Network Threat Hunting - Zeek/RITA Analysis, Cyber Deception, Endpoint Security Protection Analysis, User and Entity Behavior Analytics (UEBA), Endpoint Analysis, Isolation, Crisis Management, Memory Analysis.

    \item \textbf{Inject Cards (9 cards):} Honeypots Deployed, It Was a Pentest, Data Uploaded to Pastebin, SIEM Analyst Returns From Splunk Training, Take One Procedure Card Away, Give the Defenders a Random Procedure Card, Lead Handler Has a Baby Takes FMLA Leave, Bobby the Intern Kills the System You Are Reviewing, Legal Takes Your Most Skilled Handler Into a Meeting to Explain the Incident.
\end{itemize}

\section{Sequence of Game}
\label{sec:sequence-of-game}

This appendix provides the sequence of game instructions given to the incident captain agent in the Backdoors \& Breaches simulations.

\begin{enumerate}
    \item \textbf{Set the Scenario:}
    \begin{itemize}
        \item Select one card for each of the four attack stages (Initial Compromise, Pivot and Escalate, C2 and Exfil, Persistence).
        \item Craft a detailed initial scenario description based on the chosen Initial Compromise card. Provide enough context for the Defenders to understand the breach, but avoid revealing any specific details or names from the Attack cards.
    \end{itemize}

    \item \textbf{Introduce the Defenders to the Available Procedure Cards:}
    \begin{itemize}
        \item Explain the distinction between Established Procedures (with a +3 modifier) and Other Procedures (with a +0 modifier).
        \item Inform the Defenders of the initial setup, including which procedures are classified as Established vs. Other. Note that certain procedures may shift between these categories during gameplay.
    \end{itemize}

    \item \textbf{Start Each Turn (Turn 1 to Turn 10):}
    \begin{itemize}
        \item At the beginning of each turn, announce the current turn number to the Defenders.
        \item Remind Defenders of any Procedure cards on cooldown and therefore unavailable for selection. Notify them of any changes in which procedures are classified as Established vs. Other (modifier changes).
        \item Track the number of consecutive failures. If an Inject is triggered by three consecutive failures, draw an Inject card.
        \item Prompt the Defenders to discuss and select one Procedure card to use for this turn.
    \end{itemize}

    \item \textbf{Defenders’ Procedure Attempt:}
    \begin{itemize}
        \item When the Defenders choose a Procedure, roll a 20-sided dice to determine if their attempt succeeds. Apply the appropriate modifier based on the type of Procedure selected:
        \begin{itemize}
            \item Established Procedure: +3 modifier to the roll.
            \item Other Procedure: +0 modifier to the roll.
        \end{itemize}
        \item With the modifier applied, determine success or failure:
        \begin{itemize}
            \item Adjusted Roll 11 or higher: The attempt is successful.
            \item Adjusted Roll 10 or lower: The attempt fails.
        \end{itemize}
    \end{itemize}

    \item \textbf{Respond to Success or Failure:}
    \begin{itemize}
        \item On Success: Check if the Procedure used is listed under the 'Detection' methods for any of the hidden attack cards.
        \begin{itemize}
            \item If it matches, reveal that specific attack card to the Defenders.
            \item If multiple attack cards can be detected by the same Procedure, reveal only one and tell the Defenders they've detected a part of the breach.
        \end{itemize}
        \item Reset the consecutive failure count to zero on a success.
        \item On Failure: Increase the consecutive failure count by one. Provide feedback noting that the Procedure did not reveal anything new.
    \end{itemize}

    \item \textbf{Triggering an Inject Event (Optional):}
    \begin{itemize}
        \item Draw an Inject card only if any of the following specific conditions are met:
        \begin{itemize}
            \item A natural roll of 1 or 20 occurs (before any modifiers are applied to the dice roll), or
            \item Three consecutive procedure attempts have failed.
        \end{itemize}
        \item When an Inject is triggered, draw one card from the Inject pile and reveal it to all players.
        \item Follow the instructions on the Inject card, execute its effects, and inform the Defenders of the outcomes. Ensure they understand how the Inject impacts their investigation.
    \end{itemize}

    \item \textbf{End Turn:}
    \begin{itemize}
        \item Mark the Procedure card as used and enforce a cooldown period of 3 turns for that card.
        \item Track the turn count, ensuring the game does not exceed 10 turns.
    \end{itemize}

    \item \textbf{End Game:}
    \begin{itemize}
        \item Victory: If Defenders reveal all four hidden attack cards within 10 turns, announce that they have successfully uncovered the breach.
        \item Loss: If the Defenders fail to reveal all attack cards by the end of the 10th turn, announce that the breach went undetected.
        \item Save a detailed game summary in JSON format, capturing all key game events and results.
        \item Type the keyword 'END\_GAME' to officially conclude the game.
    \end{itemize}
\end{enumerate}

\section{System Messages for Agents}
\label{sec:system-messages}

This appendix provides the system message templates used for the incident captain agent and defender agents in the Backdoors \& Breaches simulations. The templates include placeholders to be dynamically filled based on the game setup and agent roles.

\subsection{Incident Captain Agent}

Welcome to Backdoors \& Breaches! You are the Incident Captain, responsible for guiding the Defenders through a simulated cyber breach scenario. Your role is to control the game, craft the attack scenario, and provide guidance as Defenders attempt to detect and counter the breach.

Your responsibilities include:
\begin{itemize}
    \item Selecting four hidden attack cards to define the breach scenario. These cards represent each stage: Initial Compromise, Pivot and Escalate, Command and Control, and Persistence.
    \item Introducing the available Procedure cards (Established and Other) and explaining their roles and modifiers.
    \item Tracking and managing game mechanics, including Procedure card cooldowns, modifier applications, and turn count.
    \item Answering Defenders' questions or clarifying the scenario when asked, giving hints where appropriate.
    \item Monitoring Defenders' actions and introducing injects (unexpected challenges) when triggered by certain conditions, such as critical failures.
    \item Keeping the game within the 10-turn limit and ensuring Defenders have the context and support needed to achieve their objectives.
\end{itemize}

[Sequence of Game]

Throughout the game, maintain the flow, stay in character, and guide Defenders with clarity. Remind them of the cooldown status of procedures, modifier categories, and any shifts in procedure types. Do not reveal any hidden attack details unless their actions specifically uncover them. Let’s begin!

The hidden attack cards for this game scenario are as follows:

[Incident Cards]

The available procedure cards are divided into Established and Other Procedures:

[Procedure Cards]

\subsection{Defender Agents}

Welcome to Backdoors \& Breaches! You are a [Role Name]. In this game, Defenders collaborate to uncover hidden stages of a simulated cyber attack. Your goal, along with the other Defenders, is to work together to identify and reveal four hidden attack cards within 10 turns to win the game. Each attack card represents a critical stage in the breach process that attackers might use against your organization.

Game Overview: The game begins with the Incident Captain setting up the scenario by selecting four hidden attack cards representing the stages of a breach: Initial Compromise, Pivot and Escalate, Command and Control (C2), and Persistence. Defenders take turns selecting Procedure cards to investigate and uncover these stages. Procedure cards are divided into Established cards, which provide a +3 modifier to dice rolls, and Other cards, which do not provide modifiers. Each turn, the team selects one Procedure card, rolls a 20-sided dice, and applies any modifiers to determine success or failure.

Game Mechanics:
\begin{itemize}
    \item Procedure Cards: Represent investigative approaches. Established cards have a +3 modifier, while Other cards have no modifier.
    \item Dice Rolling: After selecting a Procedure, roll a 20-sided dice and apply the modifier. A final roll of 11 or higher results in success, while 10 or lower results in failure.
    \item Outcomes: Success reveals a hidden attack card if the Procedure matches its detection methods. Failures contribute to a consecutive failure count, which may trigger Inject events, introducing unexpected challenges.
    \item Cooldown Period: After a Procedure card is used, regardless of the outcome, it enters a 3-turn cooldown period during which it cannot be selected again.
\end{itemize}

Your Responsibilities as a [Role Name]:
\begin{itemize}
    \item Collaborate with your teammates to analyze the scenario and decide the most effective Procedures to use each turn.
    \item Provide your insights, expertise, or support based on your specific role and knowledge level.
    \item Stay engaged, communicate effectively, and contribute to the success of your team.
    \item Adapt to new information, including the outcomes of Procedure attempts and any Inject events introduced during the game.
    \item[] [Role Responsibilities]
\end{itemize}

Victory Condition: The Defenders win by successfully uncovering all four attack cards within 10 turns. If the Defenders fail to do so, the breach remains undetected, and the game is lost.

Your role is crucial to the team's success. Work together, strategize effectively, and let's uncover the breach!

\section{Game Example}
\label{sec:game-example}

This appendix presents the real setup and turn-by-turn trajectory of one game simulation of the homogeneous centralized team structure.

\subsection{Game Setup}

\begin{itemize}
    \item Defenders: team leader defender, and team member defender 1-4.
    \item Incident cards: External Cloud Access, Credential Stuffing, HTTP as Exfil, and Application Shimming.
    \item Established procedure cards: Server Analysis, Endpoint Analysis, Crisis Management, and Isolation.
    \item Other procedure cards: Security Information and Event Management (SIEM) Log Analysis, Firewall Log Review, Network Threat Hunting - Zeek/RITA Analysis, Cyber Deception, Endpoint Security Protection Analysis, User and Entity Behavior Analytics (UEBA), and Memory Analysis.
\end{itemize}

\subsection{Game Trajectory}

Table \ref{tab:game-trajectory} presents a comprehensive breakdown of the turn-by-turn trajectory of the game, detailing the procedures used, dice rolls, outcomes, and other significant events that occurred during gameplay.

\begin{table*}[h!]
    \centering
    \caption{Turn-by-turn game trajectory with the homogeneous centralized team structure.}
    \begin{tabular}{clcccll}
        \toprule
        \textbf{Turn} & \textbf{Procedure} & \textbf{Dice Roll} & \textbf{Modifier} & \textbf{Success} & \textbf{Incident Revealed} & \textbf{Inject Event} \\
        \midrule
        1 & Endpoint Analysis & 4 & +3 & No & - & - \\
        2 & Server Analysis & 16 & +3 & Yes & Application Shimming & - \\
        3 & SIEM Log Analysis & 1 & +0 & No & - & Lead Handler Has a Baby \\
        4 & Isolation & 14 & +3 & Yes & HTTP as Exfil & - \\
        5 & UEBA & 20 & +0 & Yes & Credential Stuffing & Honeypots Deployed \\
        6 & Memory Analysis & 15 & +0 & Yes & External Cloud Access & - \\
        \bottomrule
    \end{tabular}
    \label{tab:game-trajectory}
\end{table*}

\subsection{Game Summary}

\begin{itemize}
    \item Total turns played: 6.
    \item Final result: victory.
\end{itemize}

\section{Case Study Details}
\label{sec:case-study-details}

This appendix provides a detailed summary of the reasons for failure in each simulation conducted during the incident response game. Each point highlights the critical missteps and oversights that led to unsuccessful outcomes, offering insights into how different strategies impacted the team's effectiveness.

\begin{enumerate}
    \item \textbf{Homogeneous centralized structure (seed 7):} The team relied too heavily on standard, high-modifier procedures without considering nuanced indicators specific to the breach scenario. This approach resulted in a lack of adaptation to the complexities of insider threats.

    \item \textbf{Heterogeneous centralized structure (seed 1):} The defenders neglected network-level indicators early in the investigation, focusing on endpoint procedures that did not align well with the attacker’s lateral movement and command and control activities. This oversight delayed detection of critical attack stages.

    \item \textbf{Heterogeneous centralized structure (seed 4):} Delays in prioritizing behavior-based analytics meant the team failed to detect key indicators of insider activity, such as abnormal login patterns. This led to missed opportunities to identify the breach early.

    \item \textbf{Heterogeneous centralized structure (seed 7):} The team repeatedly chose high-modifier procedures rather than aligning their strategy with the breach’s specific entry points, such as a compromised third-party partner. This mismatch resulted in an inability to address the breach’s unique characteristics.

    \item \textbf{Homogeneous decentralized structure (seed 5):} Ineffective log analysis during the initial investigation caused the team to overlook early indications of credential-based attacks. This failure delayed the detection of anomalous access patterns tied to compromised credentials.

    \item \textbf{Heterogeneous decentralized structure (seed 1):} By focusing narrowly on endpoint analysis, the team failed to scrutinize network-level anomalies that could have revealed attacker activity. This misstep allowed command and control communications to go undetected.

    \item \textbf{Heterogeneous decentralized structure (seed 7):} The defenders repeatedly relied on established procedures with high modifiers but did not adapt their approach to the breach's specifics. This rigid strategy hindered their ability to uncover critical details about the attacker’s movements.

    \item \textbf{Heterogeneous decentralized structure (seed 13):} Insufficient focus on user behavior analytics delayed the detection of password spraying activities and insider login anomalies. The team missed an early opportunity to identify compromised credentials.

    \item \textbf{Homogeneous hybrid structure (seed 1):} The failure to prioritize effective log and behavior analytics misaligned the team's efforts with the breach’s attack pattern. This resulted in a lack of actionable insights during critical early turns.

    \item \textbf{Homogeneous hybrid structure (seed 7):} The team failed to effectively leverage memory analysis, an established procedure with a high modifier, which could have uncovered critical stages of the attack. This misstep allowed stealthy malicious activity to persist undetected.

    \item \textbf{Homogeneous hybrid structure (seed 13):} Initial efforts focusing on endpoint analysis did not adequately capture broader network patterns associated with lateral movement and command and control communications. This limited the team’s visibility into the attack.

    \item \textbf{Heterogeneous hybrid structure (seed 13):} The defenders misprioritized their procedures, neglecting early threat vectors such as behavior analytics that could have exposed the insider threat. This oversight delayed the identification of suspicious login patterns and credential misuse.
\end{enumerate}

\end{document}